\documentclass[letterpaper, 10 pt, conference]{ieeeconf}  

\IEEEoverridecommandlockouts                              

\usepackage{graphicx} 
\usepackage{subfig}
\usepackage{epsfig} 
\usepackage{mathptmx} 
\usepackage{times} 
\usepackage{amsmath} 
\usepackage{amssymb}  
\usepackage{tipa}
\usepackage{cite}
\usepackage{array}
\usepackage{hyperref}
\usepackage{fixltx2e}
\usepackage{color}
\usepackage{dblfloatfix}

\usepackage{url}

\usepackage{tikz}
\usepackage{algorithm}
\usepackage{algorithmic}
\usepackage{booktabs}
\usepackage{float}
\usepackage{amsmath,amssymb,amsfonts}
\usepackage{mathrsfs}
\usepackage{soul,color}
\usepackage{graphicx}
\graphicspath{{figs/}}
\usepackage{lipsum}
\usepackage[utf8]{inputenc}
\usepackage{multicol,tabularx,capt-of}
\usepackage{multirow}
\usepackage{nicematrix}

\newcommand\bv[1]{{\color{black}#1}}
\newcommand\rv[1]{{\color{black}#1}}

\title{\LARGE \bf
Identifying Reaction-Aware Driving Styles of Stochastic Model Predictive Controlled Vehicles by Inverse Reinforcement Learning
}

\author{Ni~Dang$^{1}$, Tao~Shi$^{1}$, Zengjie Zhang$^{2*}$, Wanxin Jin$^{3}$, Marion Leibold$^{1}$, and~Martin~Buss$^{1}$
\thanks{The work is partially supported by the European project SymAware under the grant Nr. 101070802.}
\thanks{$^{1}$Ni~Dang, Tao~Shi, Marion Leibold and Martin Buss are with the Chair of 
Automatic Control Engineering at the Technical University of Munich,
Munich, Germany. (Email: \{\tt\small ni.dang; tao.shi; marion.leibold; mb\}@tum.de) }
\thanks{$^{2}$Zengjie Zhang is with the Department of Electrical Engineering, Eindhoven University of Technology, Eindhoven, the Netherlands. (Email: \tt\small z.zhang3@tue.nl)}
\thanks{$^{3}$Wanxin Jin is with the School for Engineering of Matter, Transport, and Energy, Arizona State University, Tempe, USA. (Email: \tt\small wanxin.jin@asu.edu)}
\thanks{*Corresponding author.}}

\begin{document}

\maketitle
\thispagestyle{empty}
\pagestyle{empty}

\begin{abstract}
The driving style of an Autonomous Vehicle (AV) refers to how it behaves and interacts with other AVs. In a multi-vehicle autonomous driving system, an AV capable of identifying the driving styles of its nearby AVs can reliably evaluate the risk of collisions and make more reasonable driving decisions. However, there has not been a consistent definition of driving styles for an AV in the literature, although it is considered that the driving style is encoded in the AV's trajectories and can be identified using Maximum Entropy Inverse Reinforcement Learning (ME-IRL) methods as a cost function. Nevertheless, an important indicator of the driving style, i.e., how an AV reacts to its nearby AVs, is not fully incorporated in the feature design of previous ME-IRL methods. In this paper, we describe the driving style as a cost function of a series of weighted features. We design additional novel features to capture the AV's reaction-aware characteristics. Then, we identify the driving styles from the demonstration trajectories generated by the Stochastic Model Predictive Control (SMPC) using a modified ME-IRL method with our newly proposed features. The proposed method is validated using MATLAB simulation and an off-the-shelf experiment.

\end{abstract}

\section{INTRODUCTION}
The driving style of an Autonomous Vehicle (AV) refers to how it generally achieves its driving goal and interacts with other vehicles, e.g., how to make driving decisions according to the current states, the desired speed, or the collision avoidance requirements~\cite{Eboli2017}.
The AV that can predict others' driving styles and incorporate the prediction into its decision-making is considered to be capable of reasonably evaluating and \textit{reacting} to the risk of collisions with other nearby AVs. A reactive AV is expected to make safer and more reasonable driving decisions than those that do not.
However, the driving style of an AV has not been consistently defined in the literature. Also, the driving styles in reactive and non-reactive situations may be different, which brings up challenges to its identification. The driving style has been represented as a cost function with weighted features~\cite{Kuderer2015, Sun2018} and can be learned from demonstration data~\cite{Abbeel2004}. 

Inverse Reinforcement Learning (IRL) that retrieves an unknown reward function from demonstration data has been widely employed to learn the driving style cost functions~\cite{Kuderer2015, Abbeel2004, Ziebart2008, Levine2012, Kretzschmar2014}. 
Among them, \cite{Abbeel2004, Ziebart2008} learn driving styles based on a stochastic Markov Decision Process (MDP), \rv{adopting a probabilistic transition model}. However, high-order properties, such as acceleration, can not be incorporated into the feature design with stochastic MDPs, although they are obviously important to determining driving styles. Different from them, \rv{in~\cite{Levine2012, Kretzschmar2014}, deterministic MDPs  are used to model the vehicle dynamic such that accelerations are considered in the feature design. Instead of modeling the dynamics as deterministic MDP, \cite{Kuderer2015}~represents trajectories using time-continuous splines that allow for incorporating acceleration into the feature design.}
IRL uses a linear combination of the features to capture the characteristics of trajectories \cite{Kuderer2012, Sun2018}. 
The IRL method aims to find the optimal weights of features and reproduce a trajectory that best mimics the driving style encoded in the demonstration trajectory generated by an expert. However, to the best of our knowledge, these methods have only been mainly applied to single-AV cases where the reactions among different AVs were rarely considered. Specifically, how an AV reacts to a nearby AV is not incorporated in the features used to identify the driving style. 

Learning the driving style using IRL methods requires demonstration data. \bv{Stochastic Model Predictive Control (SMPC) is capable of generating demonstration trajectories that encode the desired driving styles.}
The driving style of an AV controlled by SMPC depends heavily on the risk parameter in the probabilistic constraint to avoid colliding with obstacles \cite{Carvalho2014, Tim2018, Tim2021gaussian, Tim2021journal, Dang2023}. How the risk parameter qualitatively affects the driving style of an SMPC-controlled vehicle has been described in \cite{Dang2023}. A greater risk parameter leads to a more conservative driving style, and vice versa \cite{Dang2023}. Therefore, \bv{SMPC can get qualitatively aggressive or conservative driving styles by simply adjusting the risk parameter.} 
 
In this paper, we solve the driving style identification problem for a \rv{two}-vehicle system. We stand on the position of the ego AV and identify the driving styles of its \rv{nearby AV} using a Maximum Entropy IRL (ME-IRL) method. Different from the conventional methods used for single-vehicle cases~\cite{Kuderer2015, Levine2012, Kretzschmar2014},
we design four additional features to depict the ego AV's reactions to its \rv{nearby AV}.
Among them, three are active only when the AV is close to the \rv{nearby AV}. This requires a triggering condition \bv{to activate them}.
The detailed contributions are summarized as follows:
\begin{itemize}
    \item We propose four novel features to capture the reaction-aware characteristics of the driving style for a \rv{two}-vehicle case;    
    \item We design a triggering condition based on an elliptical index to \rv{activate the reaction-aware features}.
\end{itemize}

The rest of the paper is organized as follows. Sec. \ref{Preliminaries} introduces the preliminaries containing SMPC, trajectory representation, and ME-IRL. In Sec. \ref{MainMethods}, we present the modified ME-IRL method to identify the driving style from the demonstration trajectory. The simulation studies that validate the efficacy of the proposed method are shown in Sec. \ref{simulations}. Finally, Sec. \ref{conclusion} concludes the paper.

\section{Preliminaries}\label{Preliminaries}
Preliminary to the main results, we first present an SMPC formulation including the safety constraints with the risk parameter. Then, we describe how we represent trajectories. Finally, we briefly introduce the ME-IRL method.
\subsection{SMPC and Safety Constraint}\label{SMPCformulationAndConstraint}
We use SMPC to generate demonstration trajectories. The formulation of SMPC and the safety constraints are introduced as follows.
\subsubsection{SMPC Formulation}\label{mpc}
An Ego Vehicle (EV) that avoids a Target Vehicle (TV) solves the following optimal control problem at each time step \cite{Dang2023}:
\begin{subequations}
	\begin{align}
		\mathop{\min}\limits_{\pmb{u}} \ \  &J(\pmb{\xi}, \pmb{u})\label{costfunction1}\\
		\text{s. t. }\ &{\xi}_{k+1}=\mathscr{F}({\xi}_k, {u}_k), \ k=0,1,\cdots,N-1,\label{systemdynamics1}\\
		&{\xi}_{k} \in \Xi,\ k=0,1,\cdots,N,\label{statecon1}\\
		&{u}_{k} \in \mathcal U,\ k=0,1,\cdots,N-1,\label{inputcon1}\\
		&\Pr \ ({\xi}_k\in\Xi_k^{\text {safe}}) \ge p, \ k=1,2,\cdots,N,\label{probcon1}
	\end{align}	
\end{subequations}
with the state $\xi_k$ and control input $u_k$ at prediction step $k$, where $k$ counts from current time $t$ on. 
The state vector
$\xi_k = {[x_k, y_k, \phi_k, v_k]}^\intercal$ consists of longitudinal position $x_k$, lateral position $y_k$, inertial heading of the vehicle orientation $\phi_k$ and the vehicle velocity $v_k$ at prediction step $k$.
The control input vector $u_k = [a_k,\delta_k]^\intercal$ contains the acceleration $a_k$ and the front steering angle $\delta_k$ at prediction step $k$. 
The sequences $\pmb{\xi} = \{\xi_0, \xi_1, \dots, \xi_N\}$ and $\pmb{u} = \{u_0, u_1, \dots, u_{N-1}\}$ are the system states and inputs over the entire prediction horizon~$N$. 
The cost function $J$ \eqref{costfunction1}
is optimized over the control input sequence $\pmb{u}$ along the prediction horizon $N$. We expect the EV to track reference states $\boldsymbol{\xi}^{\text {ref}}_k$ and do not have big control inputs; therefore, $J=\sum_{k=0}^{N-1}({\Vert \boldsymbol{\xi}_k - \boldsymbol{\xi}_k^{\text {ref}} \Vert}_Q^2+{\Vert \boldsymbol{u}_k \Vert}_R^2)+{\Vert \boldsymbol{\xi}_N - \boldsymbol{\xi}_N^{\text {ref}} \Vert}_{Q_N}^2$. The weighting matrices are  $Q$~$\in$~$\mathbb R^{4\times4}$, $R$~$\in$~$\mathbb R^{2\times2}$ and $Q_N$~$\in$~$\mathbb R^{4\times4}$. Constraints include the system dynamic $\mathscr{F}$ of the EV in \eqref{systemdynamics1}, utilized to generate EV predictions. The sets $\Xi$ and $\mathcal U$ in \eqref{statecon1} and \eqref{inputcon1} denote the sets of admissible states and the control inputs, respectively, where the road boundaries, physical limitations of the EV and the traffic rules are taken into consideration. A safety constraint to avoid colliding with the TV is introduced by the probabilistic constraint \eqref{probcon1}, where $\Xi_k^{\text {safe}}$ depends on the prediction of the TV where the uncertainty in the prediction is taken into account by Gaussian distributions.
$\Pr (*) \ge p$ means that the hard constraint $*$ is satisfied at least with probability $p \in (1,0)$.

\subsubsection{Safety Constraint with Risk Parameter}\label{SafetyConstraint}
The safety constraint ensures that the EV remains outside \bv{a convex region} \cite{Carvalho2014} around the TV with a probability of $p$. Thus, $p$ is a risk parameter. As in \cite{Tim2018, Bruno2019, Jewison2015, Schimpe2020}, we choose the ellipse region as the convex region.
The center of the ellipse is also the center of the TV. 
Given the longitudinal distance $\Delta x_k=x_k-x_k^{\text TV}$ and the lateral distance $\Delta y_k=y_k-y_k^{\text TV}$ between the EV and TV,
the hard constraint that keeps the EV outside the ellipse region is \looseness=-1
\begin{equation}
\textstyle	d_k=\frac{{\Delta x_k}^2}{{l_a}^2}+\frac{{\Delta y_k}^2}{{l_b}^2}-1\ge0, \label{safecon}\\
\end{equation}
where the size of the safety ellipse is determined by the length of the semi-major axis $l_a$ and semi-minor axis $l_b$. $d_k \ge 0$ is one way to realize the hard constraint $\xi_k\in\Xi_k^{\text {safe}}$ in \eqref{probcon1}.
We soften constraint $d_k \ge 0$ employing a probabilistic constraint $\Pr \ (d_k\ge 0)\ge p$.
The safety constraint is active only when the EV is close to the TV and reacts to the TV.
The risk parameter dominantly influences the driving style of the EV.  The greater $p$ is, the less aggressive the EV is, and vice versa.

\subsection{Trajectory Representation}\label{QuinticPolynomials}
We represent the demonstration and the reproduced trajectories using piecewise quintic spline segments to ensure their smoothness \rv{and continuity}. \rv{The continuous-time splines allow for the existence of integral feature functions}. The spline segments
are parameterized using control points comprising the positions, velocities, and accelerations in the longitudinal and lateral directions \cite{Sprunk2011}.

\subsubsection{Control Points}
We employ piecewise quintic splines ${\bf s}_j(t)$ to represent a trajectory ${\bf r}$, i.e., ${\bf r}(t)={\bf s}_j(t)$ for $t\in [t_j,t_{j+1}]$, $j\in\{0,\cdots,S-1\}$, where $S$ denotes the number of spline segments. Spline ${\bf s}_j$ is parameterized employing the pair of control inputs ${\bf c}_j$ and ${\bf c}_{j+1}$ at interval $[t_j,t_{j+1}]$~\cite{Kuderer2015}. Control input ${\bf c}_j$ is comprised of positions, velocities and accelerations in the longitudinal and lateral directions at interval $[t_j,t_{j+1}]$~\cite{Kuderer2015},
\begin{equation}
	{\bf c}_j 
	= 
	\left[
	\begin{array}{c}
		{\bf c}_j^x\\
		{\bf c}_j^y\\
	\end{array}
	\right]
	=
	\begin{bmatrix}
		[r_j^x & v_j^x & a_j^x]^\intercal\\
		[r_j^y & v_j^y & a_j^y]^\intercal\\
	\end{bmatrix},
\end{equation}
where $r^x_j$, $v^x_j$ and $a^x_j$ are the position, velocity, and acceleration of the vehicle in the longitudinal direction, respectively; $r^y_j$, $v^y_j$ and $a^y_j$ are the lateral counterparts. All control points ${\bf c}_j$ at any time $t_j$ constitute the set of control points $\bar{\bf c}=[{\bf c}_0^{\top} 
\ {\bf c}_1^{\top} \ \cdots \ {\bf c}_S^{\top}]^{\top} \in \mathbb{R}^{6 (S+1)}$. The control point ${\bf c}_0$ is fixed during the learning process. How we get control points for demonstration trajectories and reproduced trajectories will be introduced in Sec. \ref{GeneratingControlPoints}.

\subsubsection{Quintic Polynomials}
To define a 2D quintic polynomial \cite{Simon1999} of spline 
\begin{equation}
\begin{array}{c}
{\bf s}_j(\tau)
	=
	\left[
	\begin{array}{c}
		q_5^x\tau^5+q_4^x\tau^4+\cdots+q_1^x\tau+q_0^x \\
		q_5^y\tau^5+q_4^y\tau^4+\cdots+q_1^y\tau+q_0^y  \\
	\end{array}
	\right]
	\end{array}
\end{equation}
where $\tau \in [t_j,t_{j+1}]$,
twelve coefficients $q_5^x,\cdots,q_0^x,q_5^y,\cdots,q_0^y$ are required. 
The coefficients $q_5^x,\cdots,q_0^x$ for the longitudinal direction can be obtained using control points ${\bf c}_j^x=[r^x_0,v^x_0,a^x_0]$ at time $t_j$ and ${\bf c}_{j+1}^x=[r^x_{T},v^x_{T},a^x_{T}]$ at time $t_{j+1}$, as shown in equations (7) and (8) in \cite{Simon1999}.
If we replace the longitudinal part of the control points with their lateral parts, we obtain coefficients $q_5^y,\cdots,q_0^y$. 
For spline ${\bf s}_{j+1}$, we use control points $[{\bf c}_{j+1},{\bf c}_{j+2}]$ to calculate the coefficients. Since two adjacent spline segments share control points, we have continuous velocity ${\bf v}(t)$ and acceleration ${\bf a}(t)$ along the entire trajectory. 
Given the above, the set of all control points $\bar{\bf c}$ parameterize the trajectory ${\bf r}$ consisting of $S$ segments.

\subsection{ME-IRL}\label{MaximumEntropyIRL}
The driving style is quantified by a cost function represented by a linear combination of features that capture the important characteristics of the trajectories.
The ME-IRL method aims to identify the weights of the features that best fit the driving style of the demonstration trajectory and reproduce trajectories that mimic the driving style of the demonstration.~\cite{Kuderer2015, Levine2012, Kretzschmar2014}.
The features together with their weights describe the driving style and can be used to measure the similarity between trajectories.

\subsubsection{Features}\label{generalfeatures}

We adopt the following six features for an individual AV from~\cite{Kuderer2015} and~\cite{Wu2020}. These features include accelerating, approaching or maintaining desired speed, and approaching or remaining in the target lane \cite{Kuderer2015}. 
\begin{itemize}
	\item[a)] $x$-acceleration: $f_{\mathrm{ax}}=\int_t{\left\Vert\ddot{r}^x(\tau)\right\Vert}^2 \mathrm{d} \tau$. 
	\item[b)] $y$-acceleration: $f_{\mathrm{ay}}=\int_t{\left\Vert\ddot{r}^y(\tau)\right\Vert}^2 \mathrm{d} \tau$. 
    \item[c)] Desired velocity: $f_{\mathrm{v}}=\int_t{\left\Vert v^x_{\text {des}}- \dot{r}^x(\tau)\right\Vert}^2 \mathrm{d} \tau$, where $v^x_{\text {des}}$ is the desired velocity in the longitudinal direction. 
	\item[d)] Desired lane: $f_{\mathrm{l}}=\int_t{\left\vert l_{\text{des}} - r^y(\tau)\right\vert} \mathrm{d} \tau$, where $l_{\mathrm{des}}$ is the desired lane.	
	\item[e)] Initial lane: $f_{\mathrm{il}}=\int_0^{t_{\text{turn}}}{\left\vert l_{\text{initial}} - r^y(\tau)\right\vert} \mathrm{d} \tau$, where $l_{\mathrm{initial}}$ is the initial lane of the EV and $t_{\text{turn}}$ is the time when the EV remains in $l_{\mathrm{initial}}$. 
	\item[f)] End lane: $f_{\mathrm{el}}=\int_{t_\text{end-1}}^{t_\text{end}}{\left\vert l_{\text{target}} - r^y(\tau)\right\vert} \mathrm{d} \tau$, where $l_{\text{target}}$ is the target lane of the EV at the ending time $t_{\mathrm{end}}$.
\end{itemize}

Therefore, the features can be recognized as functions of a trajectory ${\bf r}$. All features are collected in a feature vector $\bf{f}(\bf{r})$, which will be introduced in Sec.~\ref{MainMethods}.

\subsubsection{Learning of Weights}

The weights of the features are summarized in a weight vector ${\pmb{\theta}} = [\theta_1,\cdots,\theta_m]^\intercal \in \mathbb{R}^m$. Then, a trajectory can be reproduced by a learned weight vector ${\pmb{\theta}}$, denoted as ${\bf r}_{\pmb{\theta}}$.
We aim to find the optimal feature weight ${\pmb{\theta}}^*$ such that the features of the reproduced trajectory ${\bf r}_{\pmb{\theta}^*}$ are closest to those of the demonstration trajectory ${\bf r}_{\mathcal{D}}$, i.e.,
\begin{equation}	
\textstyle {\pmb{\theta}}^*=\mathop{\arg\min}_{\pmb{\theta}} \epsilon_{\pmb{\theta}} \label{ulcf}
\end{equation}
where $\epsilon_{\pmb{\theta}} = \left\Vert  \bf{f}(\bf{r}^*_{\pmb{\theta}})  - {\bf f}({\bf r}_{\mathcal{D}}) \right\Vert_2$ is the learning error, $\bf{r}^*_{\pmb{\theta}} = {{\arg\min}_{\bf{r}}} $$L(\pmb{\theta}, \bf{r})$, where $L(\pmb{\theta}, \bf{r}) = {\pmb{\theta}}^\intercal{\bf f}(\bf r)$ is a cost function that represents the driving style. This is a bilevel optimization problem~\cite{colson2007overview}. The optimal weight $\pmb{\theta}$ can be solved using an updating law ${\pmb{\theta}}={\pmb{\theta}}+\alpha\nabla_{\pmb{\theta}}$ with a learning rate~$\alpha$, where $\nabla_{\pmb{\theta}}=\mathbb{E}_{p({\bf r}\mid {\pmb{\theta}})}[{\bf f}({\bf r})]-{\bf f}(\bf{r}_{\mathcal{D}})$, where $\mathbb{E}_{p({\bf r}\mid {\pmb{\theta}})}[{\bf f}({\bf r})]\approx{\bf f}\left(\bf{r}^*_{\pmb{\theta}}\right)$, according to~\cite{Kuderer2012}. The learning is conducted in an iterative manner, i.e., the weights $\pmb{\theta}$ are used to derive a trajectory $\bf{r}_{\pmb{\theta}}$ which is then used to produce the gradient $\nabla_{\pmb{\theta}}$ for the update of $\pmb{\theta}$. The iteration of the learning process terminates when the incremental learning error is smaller than a given threshold, i.e., $\left\vert \epsilon_{\pmb{\theta}}^i - \epsilon_{\pmb{\theta}}^{i-1} \right\vert < \bar{\epsilon}$, where $\epsilon_{\pmb{\theta}}^i$ is the learning error at iteration $i\in \mathbb{N}^+$ and $\bar{\epsilon} \in \mathbb{R}^+$ is a predefined threshold.

\section{Modified ME-IRL}\label{MainMethods}
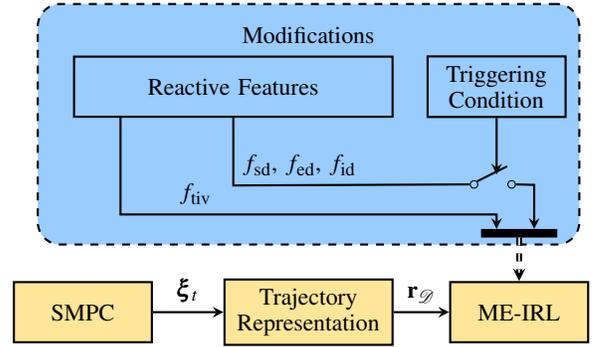
\begin{figure}[htbp]
\centering
\noindent

\begin{tikzpicture}[scale=1,font=\small]

\def\nw{2cm}
\def\nh{0.8cm}

\definecolor{s_pink}{RGB}{255, 153, 153}
\definecolor{s_blue}{RGB}{153, 204, 255}
\definecolor{s_yellow}{RGB}{255, 230, 153}

\node[minimum height=4*\nh,minimum width=3.6*\nw, text width=2.5*\nw,align=center,draw,dashed, thick,fill=s_blue, rounded corners=0.3cm] (bk) at (0cm,4.5cm) {};
\node[align=center, anchor=north] (bkn) at ([yshift=-0.2cm] bk.north) {Modifications};
\node[minimum height=\nh,minimum width=0.8*\nw, text width =0.8*\nw,align=center,draw,thick,fill=s_yellow] (smpc) at (-3cm,2cm) {SMPC};
\node[minimum height=\nh,minimum width=\nw,text width =\nw,align=center,draw,thick,fill=s_yellow] (t_rep) at (0cm,2cm) {Trajectory Representation};

\node[minimum height=\nh,minimum width=0.8*\nw,text width =0.8*\nw,align=center,draw,thick,fill=s_yellow] (irl) at (2.8cm,2cm) {ME-IRL};

\draw[->,>=stealth,thick] (smpc.east) -- node[pos=0.5,align=center, anchor=south]{$\pmb{\xi}_t$} (t_rep.west);
\draw[->,>=stealth,thick] (t_rep.east) -- node[pos=0.5,align=center, anchor=south]{${\bf r}_{\mathcal{D}}$} (irl.west);

\node[minimum height=\nh,minimum width=0.8*\nw,text width =0.8*\nw,align=center] (safety) at (-2.5cm,5cm) {};
\node[minimum height=\nh,minimum width=0.8*\nw,text width =0.8*\nw,align=center,draw,thick] (trigger) at (2.5cm,5cm) {Triggering Condition};
\node[minimum height=\nh,minimum width=0.8*\nw,text width =0.8*\nw,align=center] (old_react) at (0cm,5cm) {};
\node[minimum height=\nh,minimum width=2*\nw,text width =2*\nw,align=center,draw,thick] (react) at (-1cm,5cm) {Reactive Features};

\node[circle,inner sep=0pt,minimum size=1mm,draw] (swr) at (2.7cm, 3.7cm) {};
\node[circle,inner sep=0pt,minimum size=1mm,draw] (swl) at ([xshift=-0.5cm] swr.center) {};

\node[minimum height=0.1cm,minimum width=1cm,inner sep=0pt,draw,fill=black] (bar) at (2.8cm, 3.05cm) {};
\draw[thick] (react.south) -- ([yshift=-0.9cm] react.south) -- node[pos=0,align=left, anchor=south west]{$f_{\mathrm{sd}}$, $f_{\mathrm{ed}}$, $f_{\mathrm{id}}$} (swl.west);
\draw[thick] (swl.north east) -- ([xshift=0.4cm, yshift=0.2cm] swl.north east);

\draw[->,>=stealth,thick] (trigger.south) -- ([yshift=-0.3cm] trigger.south) -- ([yshift=-0.3cm] trigger.south) -- ([yshift=-0.75cm] trigger.south);
\draw[->,>=stealth,thick] (swr.east) -- ([xshift=0.2cm, yshift=1.28cm] irl.north) -- ([xshift=0.2cm] bar.north);

\draw[->,>=stealth,thick] (safety.south) -- ([yshift=-1.7cm] safety.center) -- node[pos=0.2,align=center, anchor=south]{$f_{\mathrm{tiv}}$}([xshift=-0.3cm, yshift=1.3cm] irl.center) -- ([xshift=-0.3cm] bar.north);

\draw[->,>=stealth,thick,double,dashed] (bar.south) -- (irl.north);

\end{tikzpicture}
\caption{The overall framework of the methods}
\label{fig:framework}
\end{figure}

To identify the driving style of an AV in a \rv{two}-vehicle system where \rv{the ego AV} reacts to the \rv{other's behaviors}, we modify the ME-IRL method in~\cite{Kuderer2015, Levine2012, Kretzschmar2014}. 
The main modifications include four novel reactive features that capture the characteristics while the AV is reacting to \rv{its nearby AV}. One of them ($f_{\mathrm{tiv}}$) is active during the entire time and the other three ($f_{\mathrm{sd}}$, $f_{\mathrm{ed}}$, $f_{\mathrm{id}}$) are triggered only when the AV is close to the \rv{nearby AV}.
Additionally, the demonstration trajectory ${\bf r}_{\mathcal{D}}$ is generated in the form of states $\pmb{\xi}_t$ by SMPC and then re-represented using Quintic Polynomials.
The overall framework of our method is shown in Fig. \ref{fig:framework}.

\subsection{Novel Features}\label{features}

For a two-vehicle system that contains an EV and a TV, we propose four novel features to capture the reaction of the EV to the TV when it tries to avoid collisions, apart from the existing features introduced in Sec. \ref{generalfeatures}.
\begin{itemize}
	\item[g)] The inter-vehicular time (TIV)-based feature: $f_{\mathrm{tiv}}=\int_{t} \frac{v^x_{\text{lane}}}{\left\vert r^x_{\text{TV}}(\tau)-r^x_{\text{EV}}(\tau)\right\vert} \mathrm{d} \tau$, where $v^x_{\text{lane}}$ is the limit velocity of the target lane, which is also the desired speed of the target lane.
	\item[h)] Start distance: $f_{\mathrm{sd}} = e^{-\left\vert r_\text{EV}^y(t_{\text{trg}})-r_\text{TV}^y(t_{\text{trg}}) \right\vert}$, where $t_{\text{trg}}$ is determined by the triggering condition which will be discussed in Sec.~\ref{featuretrigger}. 
	\item[i)] End distance:~$f_{\mathrm{ed}} \!=\! e^{\!- | {r_\text{EV}^y}({t_{\text{trg}}}\!+\!{T_\text{rct}})\!-\!r_\text{TV}^y(t_{\text{trg}}\!+\!T_\text{rct})\!|}$, where $T_\text{rct}$ is a value determined empirically.
	\item[j)] Integral distance: $f_{\mathrm{id}}=\int_{t_{\text{trg}}}^{t_{\text{trg}}+T_\text{rct}}{\left\vert r^y_\text{EV}(\tau)-r^y_\text{EV}(t_{\text{trg}})\right\vert} \mathrm{d} \tau$ which is used to capture the changes of the position in the lateral direction during time period $[t_{\text{trg}},t_{\text{trg}}\!+\!T_\text{rct}]$.
\end{itemize}

Here, the TIV-based feature $f_{\mathrm{tiv}}$ is inspired by the TIV safety metric which is commonly used to specify the collision-avoidance requirements~\cite{Dang2022}. In general, a greater TIV stands for a safer situation. We require a smaller cost for a safer situation. Therefore, We take the reciprocal of TIV as a feature.
The distance-based features $f_{\mathrm{sd}}$, $f_{\mathrm{ed}}$, and $f_{\mathrm{id}}$ are proposed to capture the characteristics between the EV and TV in the lateral direction from the triggered time $t_{\mathrm{trg}}$ to the end of the reaction. They are important for collision avoidance between two vehicles. Note that only the TIV-based feature $f_{\mathrm{tiv}}$ is active through the entire trajectories. The rest features, namely $f_{\mathrm{sd}}$, $f_{\mathrm{ed}}$, and $f_{\mathrm{id}}$ are only valid when the two vehicles are sufficiently close to each other with the concern of collisions raised. As shown in Fig.~\ref{fig:framework}, a triggering condition is set to activate these three features for ME-IRL, which will be interpreted in the following subsection.

\subsection{Triggering Condition}\label{featuretrigger}

The triggering condition for the reactive features $f_{\mathrm{sd}}$, $f_{\mathrm{ed}}$, and $f_{\mathrm{id}}$ is based on an elliptical index used to describe the squared elliptical distance \cite{Karafyllis2021} between the positions of the EV and the TV, which allows for a more accurate approximation of the physical dimensions of a vehicle, i.e.,
\begin{equation}
\textstyle	s_e=\frac{{\Delta x_t}^2}{{l_a}^2}+\frac{{\Delta y_t}^2}{{l_b}^2}\label{scalcoe}
\end{equation}
where $\Delta x_t$ and $\Delta y_t$ are the longitudinal and lateral distances between the EV and TV at time $t$. Here, the values of parameters ${l_a}$ and ${l_b}$ are the same as those in constraint \eqref{safecon}. Therefore, $\sqrt{s_e}$ is the elliptical distance between the EV and the TV scaled by ${l_a}$ and ${l_b}$.
We set a threshold value $\lambda$ for the triggering condition $s_e<\lambda$ which corresponds to an elliptical region, referred to as a \textit{scaled ellipse}. The reactive features $f_{\mathrm{sd}}$, $f_{\mathrm{ed}}$, and $f_{\mathrm{id}}$ are activated when $s_e<\lambda$ is satisfied for the first time. This time is referred to as the triggering time $t_{\text{trg}}$. The reactive features remain active for a duration $T_\text{rct}$ and become inactive again at time $t_{\text {trigger}}+T_\text{rct}$.

\subsection{Feature Scaling} 
The features introduced in this paper are calculated over different time periods. For example, $f_{\mathrm{tiv}}$ is active during the entire time, while $f_{\mathrm{sd}}$, $f_{\mathrm{ed}}$, and $f_{\mathrm{id}}$ are only active from $t_{\mathrm{trg}}$ to $t_{\mathrm{trg}} + T_{\mathrm{rct}}$.
Therefore, the features calculated over a longer time period tend to have greater feature values and have a stronger impact on the cost function $L(\bf{\theta}, \bf{r})$ than those over a shorter time period. 
To balance the influences of features with different time periods, we scale their values \cite{Ahsan2021} using a matrix $\Omega=\text{diag}(\omega_{ax},\omega_{ay},\cdots ,\omega_{id})\in \mathbb{R}^{10\times10}$, where $\omega_{ax},\omega_{ay},\cdots ,\omega_{id}$ are empirical scaling coefficients. Thus, the features are scaled by
${\bf f}(\bf{r}) =$$ [\omega_{ax}f_{\mathrm{ax}},\omega_{ay}f_{\mathrm{ay}},\cdots ,\omega_{id}f_{\mathrm{id}}]^\intercal \in \mathbb{R}^{10}$.
\rv{In this paper, we set $\omega_{il}$, $\omega_{el}$, $\omega_{sd}$, $\omega_{ed}$, and $\omega_{id}$ as $10$ and the others (the features are calculated over the whole time period) as~$1$. }

\subsection{Generation of Control Points}\label{GeneratingControlPoints}
The original demonstration generated by SMPC is represented in discrete time using the states $\pmb{\xi}_t = [x_t, y_t, \phi_t, v_t]^\intercal$ of the EV at each sampling time $t$. We re-represent the states by a demonstration trajectory ${\bf r}_{\mathcal{D}}$ represented by piecewise quintic spline segments. The spline segments are parameterized using control points comprised of the positions, velocities, and accelerations in the
longitudinal and lateral directions. We obtain the positions directly from the state vector. Velocities $v_t^x$ and $v_t^y$ can be calculated employing $v^x_t=v_t\cos{\phi}$ and $v^y_t=v_t\sin{\phi}$, respectively.
Acceleration $a_t^x$ and $a_t^y$ are approximated by $a_t^x=\frac{v_{t+1}-v_{t-1}}{2T_s}\cos{\phi}$ 
and $a_t^y=\frac{v_{t+1}-v_{t-1}}{2T_s}\sin{\phi}$,
respectively, where $T_s$ denotes the time interval between $t-1$ and $t$.
For reproduced trajectories, we simply calculate the velocities and accelerations using the following equations: $v^x_t=\dot{r}^x_t$, $v^y_t=\dot{r}^y_t$, $a^x_t=\ddot{r}^x_t$, and $a^y_t=\ddot{r}^y_t$.

\section{Simulation Studies}\label{simulations}
We examine the efficacy of the modified ME-IRL method with our newly designed features in several simulation studies. The simulation is run on a laptop with an i7-10875H CPU under $2.30\,$GHz. The optimization problem in SMPC is solved by employing the \textit{fmincon} function embedded in the NMPC toolbox~\cite{NMPCtoolbox} in MATLAB. 

\subsection{Simulation Setup}\label{sec:ss}
\begin{figure}[h]
    \centering
    \includegraphics[width=0.67\linewidth]{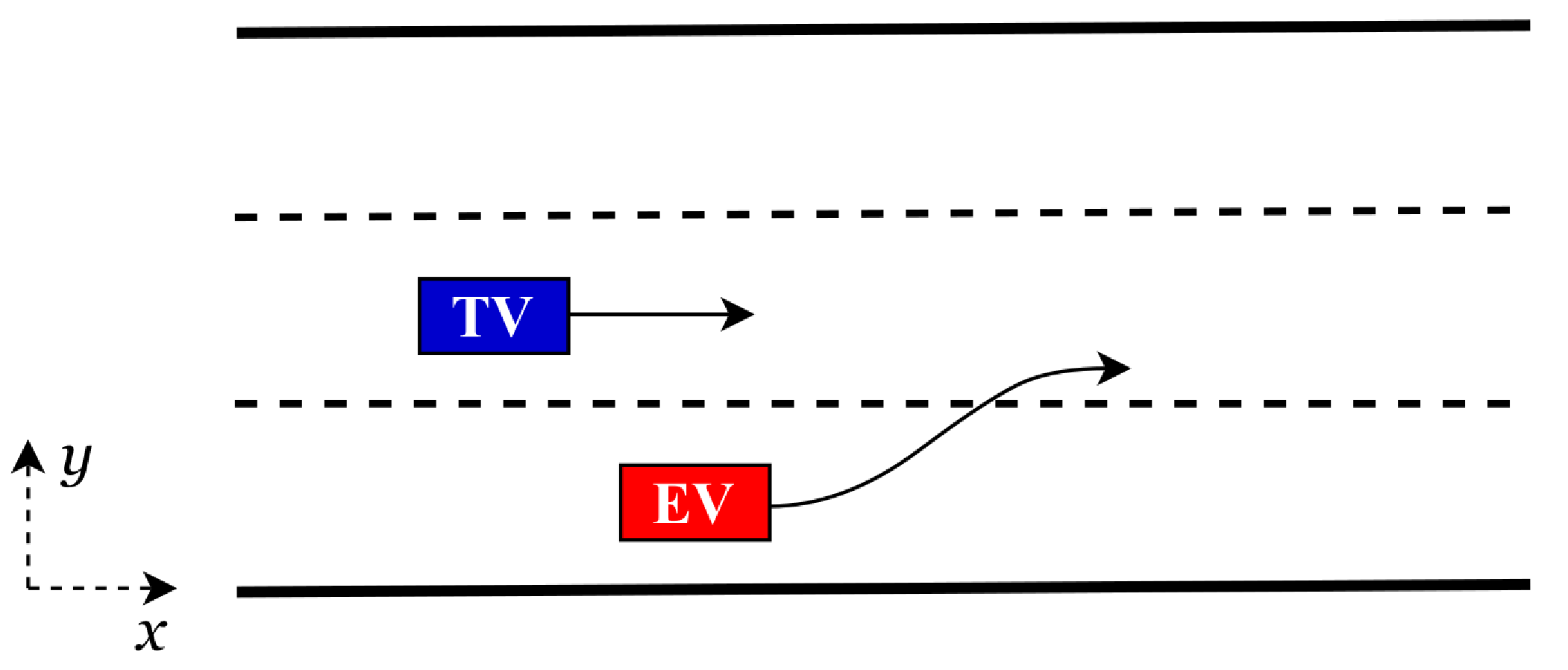}
    \vspace{-0.3cm}
    \caption{A two-vehicle lane-changing scenario.}
    \label{fig:2vehicleScenario}
\end{figure}

We consider a lane-changing scenario on a three-lane highway, as shown in Fig. \ref{fig:2vehicleScenario}. The EV (red) starts in the right lane (bottom lane) and will later move to the center lane and accelerate. The TV (blue) starts and remains in the center lane at a constant velocity $28\,$m/s. The EV tries to avoid colliding with the TV while moving toward the center lane. The initial states of the EV and TV are $\pmb{\xi}_0^\text{EV} = {[80, 2.625, 0, 25]}^\intercal$ and $\pmb{\xi}_0^\text{TV} = {[60, 7.875, 0, 28]}^\intercal$, respectively. The target velocity of the EV is $30\,$m/s. The widths of all lanes are $w^{\text{lane}}\!=\!5.25\,$m. The length and width of the vehicles are $l^{\text{veh}}\!=\!5\,$m and $w^{\text{veh}}\!=\!2\,$m, respectively. Besides, the distances from the vehicle mass center to the front axle and to the rear axle are $l_f\!=\!2\,$m and $l_r\!=\!2\,$m, respectively.

We generate the demonstration trajectories of the EV using SMPC with a risk parameter $p=0.7$, a prediction horizon $N=10$, a sampling time $T_s=0.2\,$s, and a total duration $T = 6.2\,$s (31~time steps). The boundaries in the constraints are specified as $y \in [l^{\text{veh}},3w^{\text{lane}}-l^{\text{veh}}]$, $\phi \in [-0.05,  0.05]\,$\text{rad}, $v \in [0, 70]\,$m/s, $a \in [-9, 6]\,$m/{s}$^2$ and $\delta \in [-0.05,  0.05]\,$rad. The safety ellipse is determined by the semi-major axis ${l_a}=15\,$m and semi-minor axis  ${l_b}=3\,$m.
The weighting matrices of the cost function are $Q=\text{diag}(10^{-6},0.2,50,0.2)$, $R=\text{diag}(1,10)$ and $Q_N=\text{diag}(10^{-6},0.2,50,0.2)$.
The trajectory of the TV is specified in advance. The learning termination threshold is $\bar{\epsilon} = 0.01$.

\subsection{The Demonstration Trajectories and the Trigger Time}\label{sec:dt}

\begin{figure}[htbp]
    \centering
    \includegraphics[width=0.85\linewidth]{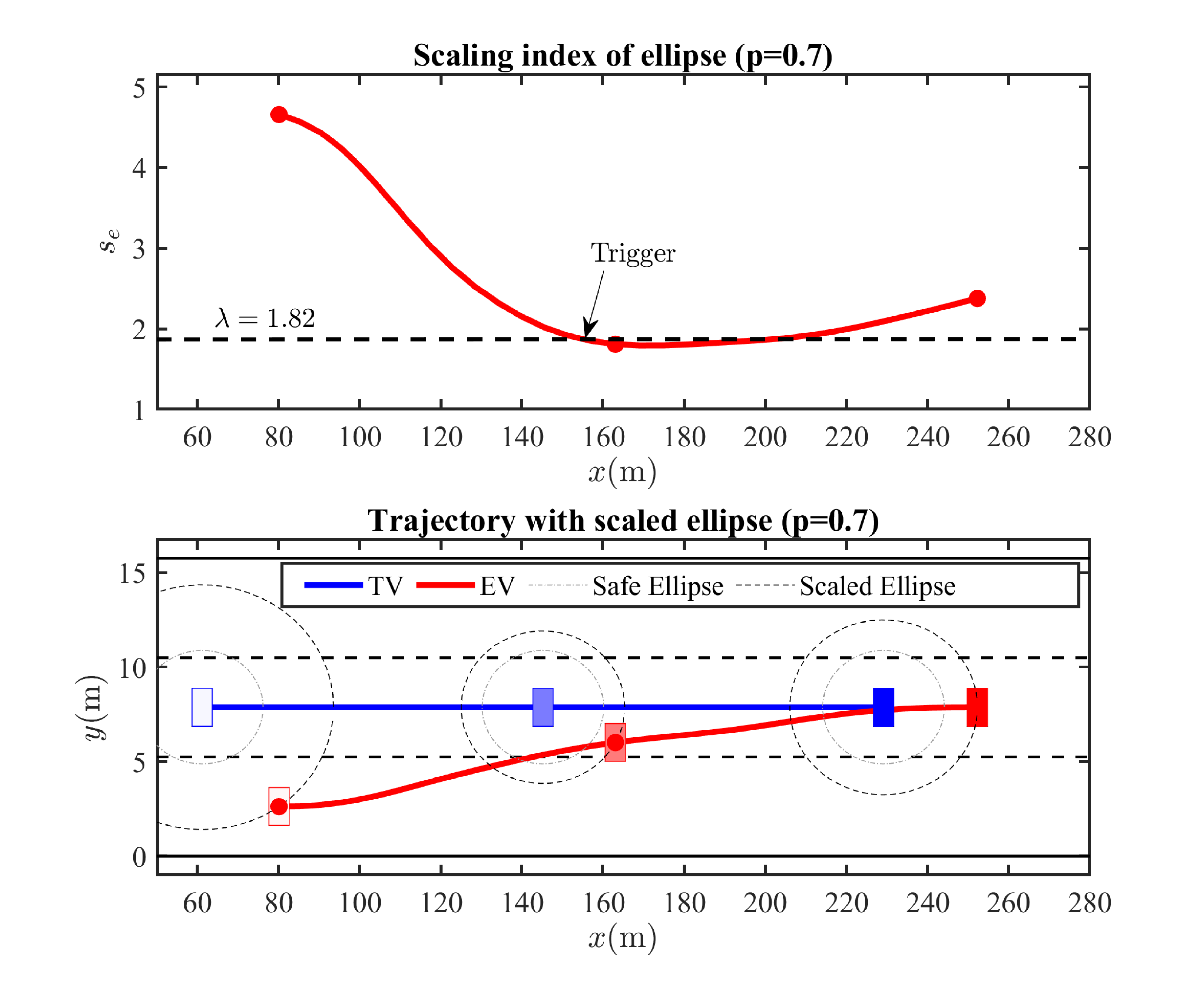}
    \caption{The elliptical index $s_e$ defined in equation~\eqref{scalcoe} and the demonstration trajectories of the EV and TV.}
    \label{fig:ScaledEllipse07}
\end{figure}

In this subsection, we show the demonstration trajectories of the EV and the TV, which reflect the reactions of the EV to the TV while avoiding potential collisions. 
We first illustrate the change of the scaling index of ellipse $s_e$ (or the elliptical index, defined in equation~\eqref{scalcoe}) along the longitudinal direction in the top subfigure of Fig.~\ref{fig:ScaledEllipse07}. The demonstration trajectories of the EV and TV with the safety and the scaled ellipses are shown in the bottom subfigure of Fig.~\ref{fig:ScaledEllipse07}, where the ellipses are only displayed at the starting, middle, and ending instants, namely the SMPC time steps 1, 16, and 31, for brevity. The positions of the vehicles at these time steps are also shown as colored rectangles. From the top subfigure, we can see that the elliptical index $s_e$ decreases from $80\,$m to $170\,$m since the two vehicles get close to each other during this period (as shown in the bottom figure). Then, from around $170\,$m, $s_e$ gradually increases since the EV actively avoids the potential collisions with the TV, which indicates the reaction of the EV to the potential collision with the TV. The EV's reaction can also be seen from the bottom subfigure, where the trajectory segment between roughly $155\,$m and $220\,$m shows a different curvature than the one before $155\,$m. 
The triggering time of the reaction can be determined as the first time when $s_e<\lambda$ is satisfied, i.e., $t_{\mathrm{trg}} = 3\,$s, where the threshold $\lambda$ is empirically set to $1.82$. 

\subsection{The Reproduced Trajectories}
\begin{figure}[htbp]
    \centering
\includegraphics[width=0.7\linewidth]{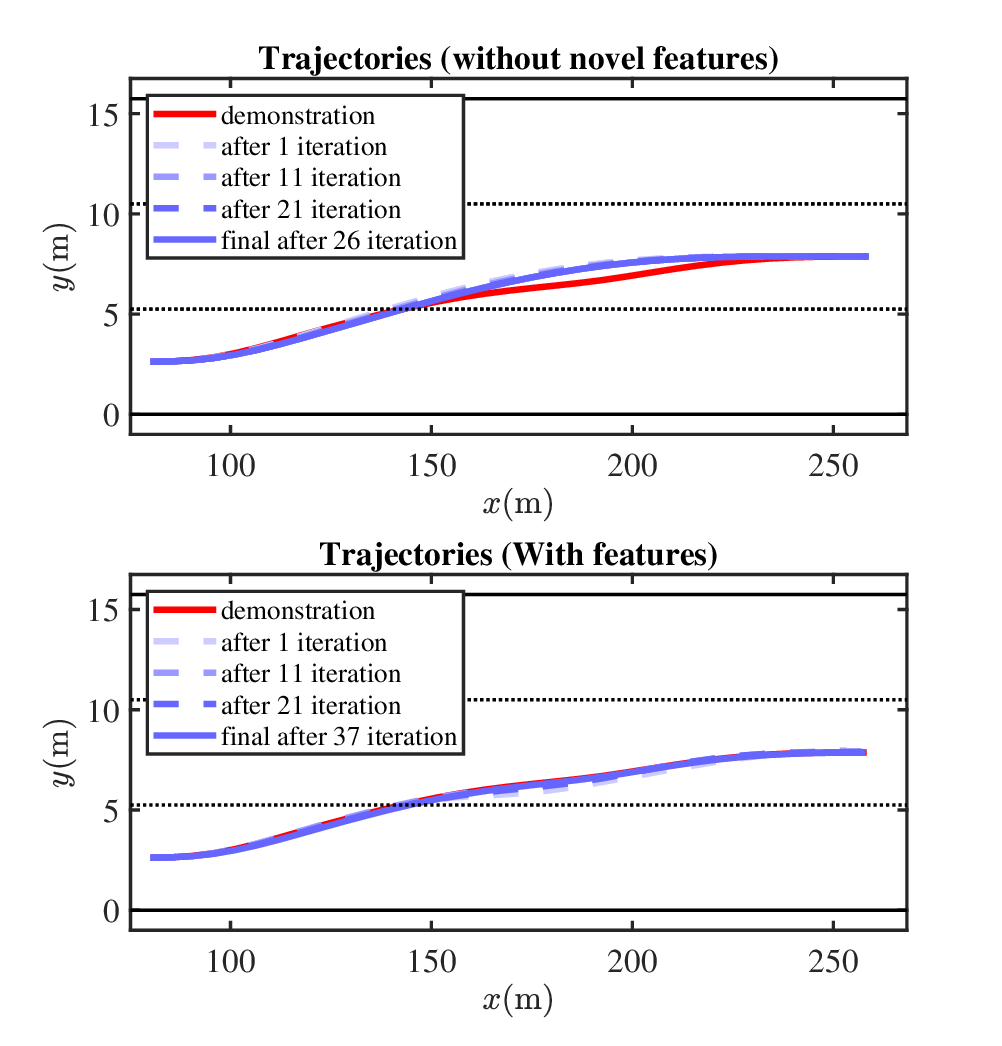}
    \caption{The reproduced trajectories of the EV using ME-IRL without (top) and with (bottom) the novel features, compared with the demonstration data.}
    \label{fig:Comparison6and10}
\end{figure}

Given the demonstration trajectories of the EV and the TV, we learn the driving style of the EV using the modified ME-IRL. Fig. \ref{fig:Comparison6and10} compares the learning performance of the ME-IRL methods without (top subfigure) and with (bottom subfigure) our novel features. Specifically, the learning performance is evaluated by the deviation between the trajectories reproduced by the learned driving styles and the demonstration trajectories.
Besides, in each subfigure, we illustrate the reproduced trajectories at iterations 1, 11, 21, and the terminal iteration. We can observe that the trajectories reproduced by the driving style with novel features better fit the demonstration trajectory, although the one without novel features is learned faster (26 over 37). Specifically, the ME-IRL without novel features produces larger trajectory gaps than the one with these features, especially between around 150 m and 220 m. This indicates that the conventional ME-IRL can not fully learn the reaction of the EV to the TV, but our modified ME-IRL can. 

\begin{figure}[htbp]
\centering
\vspace{-0.5cm}
  \includegraphics[width=0.32\textwidth]{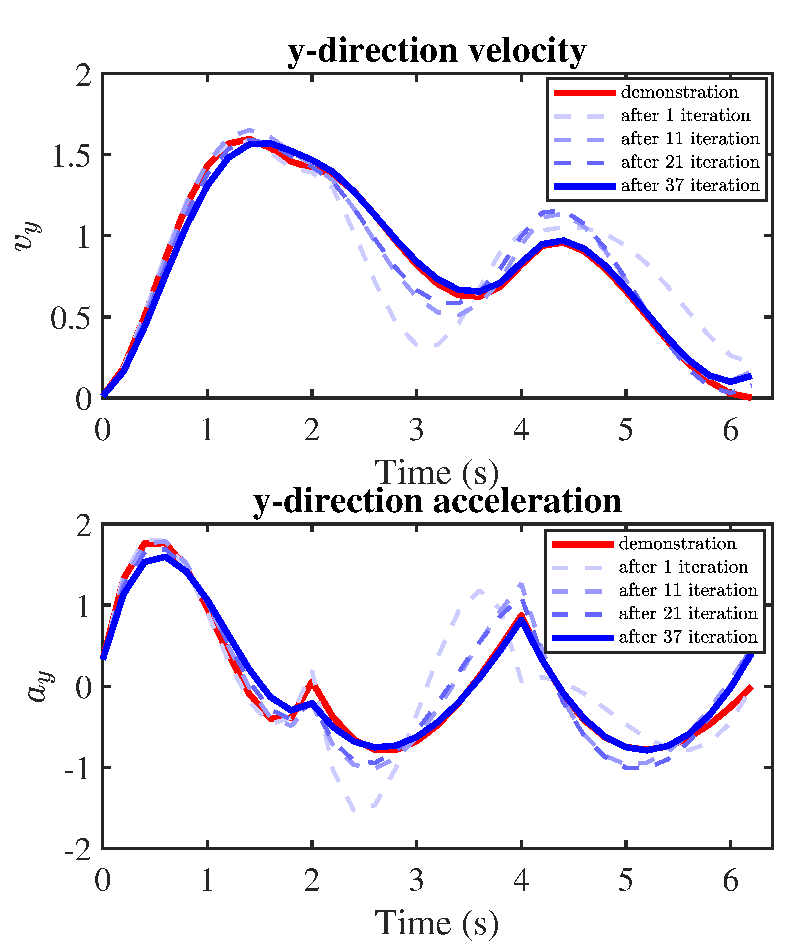}
  \caption{ The converging lateral-direction velocity and acceleration of
the EV with iterations 1, 11, 21, and 37.}
  \label{fig:VelocityAcceleration07}
\end{figure}

Similar conclusions can also be drawn from Fig. \ref{fig:VelocityAcceleration07} which displays the lateral-direction ($y$-axis) velocity and acceleration of the EV in the learning process using our modified ME-IRL method and the corresponding demonstration data. Here, we only display the lateral direction since it is much more important than the longitudinal direction for the lane-changing scenario. It can be seen that the velocity and acceleration in the lateral direction converge to those of the demonstration trajectory, which indicates successful learning. 

\subsection{Simulation experiment in Off-the-shelf Software}

To demonstrate the applicability of the proposed method to practical autonomous driving systems, we conduct an experiment in an off-the-shelf simulation environment, the Siemens\textregistered~Simcenter Prescan Software. 
The experiment setup is the same as Sec. \ref{sec:ss}. The difference is we use EV's reproduced trajectory instead of its demonstration trajectory in the experiment. This allows us to compare the trajectory reproduced using our modified ME-IRL method and the demonstration trajectory. The footage of this experimental study is published in \url{https://youtu.be/S672tUtHFyY}, where both the bird's eye view from above and the first perspective view from the TV are provided. The experimental results show that the EV's trajectory reproduced using the learned driving style with novel features is very similar to the demonstration trajectory generated in Sec.~\ref{sec:dt}. This validates the efficacy of our method. The successful experiment also demonstrates the applicability of the proposed method in practical autonomous driving systems.

\section{Conclusion}\label{conclusion}
In this paper, we extend a Maximum Entropy Inverse
Reinforcement Learning (ME-IRL) method to identify the driving styles of an Autonomous Vehicle (AV) in a \rv{two}-vehicle system incorporating the reaction among the vehicles. We propose novel features to capture the reaction-aware characteristics that indicate the driving styles of an AV while it actively avoids colliding with the \rv{nearby AV}. An elliptical index is proposed to determine the triggering time to activate some reaction-aware features. Simulation in MATLAB and experiment in \textit{Simcenter Prescan} validate the efficacy and applicability of our method. The novel features are designed for a lane-changing scenario.
In the future, we will incorporate the learned driving style into the decision-making of \rv{AVs} in \rv{multi}-vehicle lane-changing scenarios \rv{ based on real datasets, such as
INTEREATION \cite{Zhan2019}}. 

\addtolength{\textheight}{-12cm}   





\section*{ACKNOWLEDGMENT}

The license of Siemens\textregistered~Simcenter Prescan simulation software is provided through the European project SymAware under the grant Nr. 101070802. The authors would also like to thank Mr. Adem Bavar\c si and Dr. Sofie Haesaert at Eindhoven University of Technology (TU/e) for their technical support on the simulation studies of this work and thank Dr. Stephen Starck at Technical University of Munich for his advice on academic writing.

\bibliographystyle{IEEEtran}
\bibliography{Reference}

\end{document}